\documentclass[runningheads]{llncs}

\usepackage{todonotes}
\usepackage{graphicx}
\usepackage{float}
\usepackage{enumitem}
\usepackage{multicol}

\usepackage[utf8]{inputenc}
\usepackage{lmodern}
\usepackage[T1]{fontenc}
\usepackage{xparse}

\usepackage{dsfont}
\usepackage{amsfonts}
 
\usepackage[british]{babel}

\usepackage[toc,page]{appendix}
\usepackage[toc, nonumberlist, nopostdot]{glossaries}
\usepackage{glossary-longextra}
\usepackage{natbib}
\newcommand{\citepos}[1]{\citeauthor{#1}'s (\citeyear{#1})}

\usepackage[nottoc]{tocbibind}

\usepackage{caption}
\usepackage{subcaption}
\usepackage{wrapfig}

\usepackage{amssymb}
\usepackage{amsmath}
\usepackage{mathtools}

\usepackage[all]{nowidow}

\usepackage{algorithm}
\usepackage{algorithmicx}
\usepackage[noend]{algpseudocode}
\usepackage{changepage}
\usepackage{bbm}

\algrenewcommand\algorithmicindent{1.0em}

\usepackage{epigraph}
\usepackage[bottom]{footmisc}

\DeclarePairedDelimiter\brac{(}{)}
\DeclarePairedDelimiter\curlybrac{\{}{\}}
\DeclarePairedDelimiter\sqbrac{[}{]}

\renewcommand{\vec}[1]{\ensuremath{\mathbf{#1}}}

\newcommand{\Set}[1]{\ensuremath{\curlybrac*{#1}}}

\DeclareDocumentCommand \Prob {o o m} {\ensuremath{
    P \IfNoValueTF{#1}{}{_{#1}} \IfNoValueTF{#2}{}{^{#2}} \brac*{#3}
}}
\DeclareDocumentCommand \Expect {o o m} {\ensuremath{
    \mathds{E}\IfNoValueTF{#1}{}{_{#1}}\IfNoValueTF{#2}{}{^{#2}}\sqbrac*{#3}
}}
\DeclareDocumentCommand \Var {o o m} {\ensuremath{
    \mathds{V}\IfNoValueTF{#1}{}{_{#1}}\IfNoValueTF{#2}{}{^{#2}}\sqbrac{#3}
}}
\DeclareDocumentCommand \Normal {o m m} {\ensuremath{
    \mathcal{N}\IfNoValueTF{#1}{}{_{#1}}\brac{#2, #3}
}}

\newcommand{\effectiveness}{\xi}
\newcommand{\understanding}{\kappa}

\newcommand{\aemath}{\ensuremath{\text{\ae}}}

\begin{document}


\title{A Measure of Explanatory Effectiveness
\thanks{Presented at the \textit{1\textsuperscript{st} International Workshop on Trusted Automated Decision-Making (TADM) co-located with ETAPS 2021}. Work done by DC is thanks to the UKRI Centre for Doctoral Training in Safe and Trusted AI (EPSRC Project EP/S023356/1). PM wishes to thank Simon Parsons and Elizabeth Sonenberg for discussions on these topics. DC would like to thank Alex Jackson and Nandi Schoots for helping to understand understanding.}
}

\subtitle{Towards a Formal Model of Explanation}

\titlerunning{A Measure of Explanatory Effectiveness}        

\author{Dylan R. Cope\thanks{Correspondence: \url{dylan.cope@kcl.ac.uk}}         
        \and
        Peter McBurney 
}


\institute{King's College London}

\date{Received: date / Accepted: date}

\maketitle

\begin{abstract}

In most conversations about explanation and AI, the recipient of the explanation (the \textit{explainee}) is suspiciously absent, despite the problem being ultimately communicative in nature. We pose the problem `explaining AI systems' in terms of a two-player cooperative game in which each agent seeks to maximise our proposed measure of \textit{explanatory effectiveness}. This measure serves as a foundation for the automated assessment of explanations, in terms of the effects that any given action in the game has on the internal state of the explainee.

\end{abstract}

\section{Introduction}

The term \emph{explanation} in artificial intelligence (AI) is often conflated with the concepts of \emph{interpretability} and \emph{explainable AI} (XAI), but there are important distinctions to be made. \cite{Miller2019ExplanationSciences} defines interpretability and XAI as the process of building AI systems that humans can understand. In other words, by design, the AI's decision-making process is inherently transparent to a human. In contrast, explicitly explaining the decision-making to an arbitrary human is \textit{explanation generation}. The latter is the subject of this paper. More specifically, we are working towards developing a formal framework for the automated generation and assessment of explanations. 

Firstly, some key terminology: an \textit{explanation} is generated through a dialectical interaction whereby one agent, the \textit{explainer}, seeks to `explain' some phenomenon, called the \textit{explanandum}, to another agent, the \textit{explainee}. In this article, we propose a novel measure of \emph{explanatory effectiveness} that can be used to motivate artificial agents to generate good explanations (e.g. in the form of a reward signal), or to analyse the behaviours of existing communicating agents. We then define \emph{explanation games} as cooperative games where two (or more) agents seek to maximise the effectiveness measure.
\section{Related Literature}

Intepretability and XAI have received an abundance of recent attention (see \cite{Adadi2018PeekingXAI} for a review). This is largely due to two factors; regulatory demands \citep{UKInformationCommissionersOffice2019GuideGDPR} and the emergence of highly-performant black-box models, such as deep neural networks, that are naturally inscrutable. However, the central crux of interpretability techniques is the need to define a fixed \textit{interpretable domain} from which we can derive explanations. 
This presents two challenges: there are no formal procedures for determining if a given domain is interpretable; and a domain may be interpretable to some agents, but not others, or only within some contexts. 
Moving away from interpretability, the problem of explanation generation has a long history in AI \citep{Mueller2019ExplanationAI}. 
To some, there is a sense in which generating explanations is the hallmark of intelligence itself \citep{Schank1984TheGame}. To others, explanation is simply about building models -- a process which is seen as merely instrumental to intelligent behaviour \citep{Russell2010ArtificialApproach, Hutter2005UniversalIntelligence, Chaitin2006TheReason}. 


In the philosophy of science the concept of explanation is posed in terms of generating descriptions of, or hypotheses regarding, latent phenomena. This has led to investigations of formal measures of \emph{explanatory power}, with an early example being \citepos{Popper1959TheDiscovery} notion of the `degree of corroboration'. This developed into a line of philosophers devising subjectivist definitions for capturing aspects of the `goodness' of explanations or hypotheses \citep{Lipton2003InferenceExplanation, Glass2002CoherenceNetworks, Okasha2000VanExplanation, Schupbach2011ThePower}. However, by the subjectivity of these measures they may only assess the degree to which one believes (or simply likes) an explanation, which is not necessarily correlated with the degree to which an explanation is actually true (or representative of the world).


Recently, calls have been made for the need for human-centred explanation \citep{Kirsch2017ExplainAI, Abdul2018TrendsAgenda}. Yet, the framing of explanation generation as a cooperative problem between a human and machine dates back to the era of expert systems \citep{Karsenty1995CooperativeExplanation, Johnson1993ExplanationSystems, Graesser1996Question-drivenReasoning}. 
By articulating explanation as a formal dialogue, a related direction of investigation is \textit{dialogue games} \citep{McBurney2002GamesAgents}. In particular, \emph{information-seeking} \citep{Walton1995CommitmentReasoning} and \emph{education} \citep{Sklar2004TowardsEducation} dialogues are especially relevant. \cite{Sklar2018ExplanationArgumentation} conducted empirical research with a human-machine collaboration task where the agents participated in a dialogue and explanations were provided to a human based of an \textit{argumentation framework} \citep{Dung1995OnGames}.

\section{What is Explanation?} \label{sec:what_is_explanation}



In this work we treat explanatory processes as involving two agents --- an explainer and an explainee  --- and the result is that the explainee \textit{understands} the explanandum better by the end than they did at the start. We define `an explanation' as any sequence of observations made by the explainee that leads to this result. Thus an explanation could be a piece of text or spoken language, but it could also be a diagram or a piece of interactive media. 

With this we shift the problem onto formally defining a measure of an agent's `understanding' of some arbitrary phenomenon. 
We approach the question in terms of four stances\footnote{These stances do not represent arguments defended by anyone in particular, but rather we are constructing them here as rhetorical tools to help decompose the problem.} towards comprehension, understanding as: (1) a \textit{sensation} \citep{Hume1751AnUnderstanding}; (2) \textit{information compression} \citep{Chaitin2006TheReason, Zenil2019CompressionWorld, Maguire2016UnderstandingCompression}; (3) \textit{performance capacity} \citep{Turing1950ComputingIntelligence,Perkins1993WhatUnderstanding}; or (4) \textit{organised information} \citep{Lakoff1980MetaphorsBy, Hofstadter2012SurfacesThinking}.

The sensation stance states that comprehension is a conscious experience --- you understand something if you feel that you apprehend it. The compression stance says that understanding is the formulation of concise and accurate descriptions of phenomena. The performance stance argues that having information is not enough; you must also know how to use the information. The organised-information stance tells us that utilisation and compression are a byproduct of something more important; namely that the agent represents information in relation to their own conceptual framework. 
While each of the stances has issues of their own, combined they provide a persuasive account. In other words, if someone claims they understand something, they can use their information to do things, and their description of the phenomenon is concise, accurate, and grounded in other concepts that they understand, then it is hard to argue that they do not grasp the phenomenon.

\section{Technical Background}

\subsection{Algorithmic Information Theory} \label{sec:ait}




\textit{Algorithmic Information Theory} (AIT) is a view of information that takes a fundamentally computational approach \citep{Solomonoff1964A1, Kolmogorov1968ThreeInformation, Chaitin1975ATheory}. Formally, AIT is built on the notion of \emph{Kolmogorov complexity}, denoted $K(x)$. $K(x)$ is defined as the length of the shortest program, $p$, on a Universal Turing Machine (UTM), $U$, that outputs $x$.
\begin{align} \label{eqn:kolmogorov_complexity}
    K(x) = \min_p \curlybrac*{|p| ~:~ U(p) = x}, ~\text{where $|p|$ measures the length of $p$}
\end{align}
The conditional Kolmogorov complexity, $K(x|y)$, is similarly defined by the length of the shortest program that produces $x$ when given input $y$. 
\begin{align} \label{eqn:cond_kolmogorov_complexity}
    K(x|y) = \min_p \curlybrac*{|p| ~:~ U(yp) = x}
\end{align}
Thus we can define a measure of mutual information:
\begin{align} \label{eqn:kolmogorov_mutual_information}
    I(x;y) = K(y) - K(y|x)
\end{align}
Unless otherwise specified, when we talk of the mutual information between two objects we will be referring to an application of Equation \ref{eqn:kolmogorov_mutual_information}.

\subsection{Agents}

In its most basic conception, `an agent' is any system that makes observations and takes actions. For any agent $X^t \in \mathcal{X}$ at time $t$, we will denote that they make observations $o_X^t \in \mathcal{O}_X$ and take actions $a_X^t \in \mathcal{A}_X$. Another important factor in describing agents is their \textit{internal state}. This phrase can refer to various aspects of an agent's cognition, but we are mostly interested in this object insofar as it stores information. Firstly, we will assume that an agent's internal state may fall into a variety of configurations, i.e. there exists a set of possible internal states for an agent, which we will denote $\mathcal{Z}_X$. Secondly, we will talk of information being `encoded' in an agent's internal state. Given an object $o$, we will denote $X$'s encoding of $o$ as $\langle o \rangle_X$, where $\langle o \rangle_X \in \Set{p ~:~ U(p) = o}$ for some UTM $U$. We will speak of the agent `having' this encoding, or its internal state `containing' this encoding. This is independent to how this is achieved, e.g. the agent's internal state may simply store a list of encodings, or multiple encodings may overlap in a distributed storage medium such as a neural network.

\subsection{Universal Intelligence Theory} \label{sec:uit}

\textit{Universal Intelligence Theory} (UIT), proposed by \cite{Legg2007UniversalIntelligence}, establishes a definition of machine intelligence based on algorithmic information theory and reinforcement learning. In order to meaningfully compare different performances over a potentially infinite number of time steps, the scope of possible environments is limited such that the sum of rewards (the return) is always less than one. We will refer this as the set of \textit{bounded-test environments}. With this, the \textit{universal intelligence} of an agent $\pi$ is given by: 
\begin{align} \label{eqn:universal_intelligence_measure}
    \Upsilon(\pi) = \sum_{\mu \in E} 2^{-K(\mu)} V_\mu^\pi
\end{align}
Where $V_\mu^\pi$ is the return that $\pi$ achieves in environment $\mu$.

\subsection{Universal Artificial Intelligence} \label{sec:solomonoff_aixi}

Consider a stochastic environment with dynamics described by a probability distribution $\mu(e_k | \aemath_{<k})$, where $e_k$ is the percept (observation-reward tuple) given at time $k$, and $\aemath_{<k}$ is the action-percept history. In order to perform optimally, the agent in this environment must infer $\mu$. This is known as the problem of induction. By combining Solomonoff induction \citep{Solomonoff1964A1} with von Neumann-Morgenstern rational decision-making \citep{Morgenstern1953TheoryBehavior}, \cite{Hutter2005UniversalIntelligence} defines AIXI; an agent that chooses the best possible action at every time step given perfect inductive inference. 


\section{Formalising Understanding}

\subsection{Partitioning the Internal State}

In order to devise a measure of understanding, we will start by defining partitions of the information in the internal state. These partitions are constructed with respect to a given phenomenon $p \in \mathcal{P}$. There are four: the $p$-relevant information (all information related to $p$), the $p$-irrelevant (all information completely unrelated to $p$), the $p$-specific (the information that \textit{only} relates to $p$), and the $p$-background information (all information that is not specifically related to $p$). In the following formal definitions we are using a particular notation that warrants explanation. As we have already established, $z_X$ denotes the internal state of agent $X$. We denote $p$-relevant notation with a comma after the $X$ followed by $*p$, $z_{X,*p}$. The star indicates that we are `selecting' \textit{all} of the information relevant to $p$, rather than only the information specific to $p$. When the star is omitted we are referring to to specific information regarding whatever follows the comma, e.g. $z_{X,p}$ is the $p$-specific information and $z_{X,\neg p}$ is the information specific to everything that is not $p$ (the $p$-irrelevant information). 

\begin{definition}[$p$-Relevant Information]
    Given a phenomenon $p \in \mathcal{P}$ and an agent $X$ with internal state $z_X$, the $p$-relevant information $z_{X,*p} \in \mathcal{Z}_X$ is the object where $I(z_{X, *p};p) = I(z_X;p)$ and $I(z_{X,*p}; z_X)$ is minimised, i.e. there exists no $z_{X,*p}'$ such that $I(z_{X, *p}';p) = I(z_X;p)$ and $I(z_{X, *p}';z_X) < I(z_{X, *p};z_X)$.
\end{definition}


\begin{definition}[$p$-Irrelevant Information]
    Given a phenomenon $p \in \mathcal{P}$ and an agent $X$ with internal state $z_X \in \mathcal{Z}_X$, the $p$-irrelevant information $z_{X,\neg p}$ is the object where $I(z_{X, \neg p};p) = 0$ and $I(z_{X,\neg p}; z_X)$ is maximised, i.e. there exists no $z_{X,\neg p}'$ such that $I(z_{X, \neg p}';p) = 0$ and $I(z_{X, \neg p}';z_X) > I(z_{X, \neg p};z_X)$.
\end{definition}


\begin{definition}[$p$-Specific Information]
    Given a phenomenon $p \in \mathcal{P}$ and an agent $X$ with internal state $z_X \in \mathcal{Z}_X$, the $p$-specific information $z_{X,p}$ is the object where $I(z_{X,p}; p) > 0,~ I(z_{X,p}; p') = 0 ~\forall p' \in \mathcal{P}, p' \neq p $ and the mutual information $I(z_{X,p};z_X)$ is maximised, i.e. there exists no $z_{X,p}'$ such that $I(z_{X,p}'; p) > 0,~ I(z_{X,p}', p') = 0 ~\forall p' \in \mathcal{P}, p' \neq p$ and $I(z_{X,p}';z_X) > I(z_{X,p};z_X)$.
\end{definition}

\begin{definition}[$p$-Background Information]
    Given a phenomenon $p \in \mathcal{P}$ and an agent $X$ with internal state $z_X \in \mathcal{Z}_X$ and $p$-specific information $z_{X,p}$, the $p$-background information\footnote{By the notation $*\neg p$ we can see that this is `everything relevant' to `not $p$'.} $z_{X,*\neg p}$ is the object where $I(z_{X,p}; z_{X,*\neg p}) = 0$ and $I(z_{X,*\neg p}; z_X)$ is maximised, i.e. there exists no $z_{X,*\neg p}'$ such that $I(z_{X,p}; z_{X,*\neg p}') = 0$ and $I(z_{X,*\neg p}';z_X) > I(z_{X,*\neg p};z_X)$.
\end{definition}

\subsection{Information Compression}

With these partitions we can define how compressed the $p$-relevant information is:

\begin{definition}[$p$-Compression Factor]
    Suppose a phenomenon $p \in \mathcal{P}$ and an agent $X$. The $p$-compression factor $c: \mathcal{X} \times \mathcal{P} \rightarrow (0, 1]$ is given as the ratio of the Kolmogorov complexity of the $p$-relevant information object to the size of the agent's encoding of that information:
    \begin{align}
        c(X, p) = \frac{K(z_{X,*p})}{|\langle z_{X,*p} \rangle_X|}
    \end{align}
\end{definition}


\subsection{Information Utilisation}

Next, we will attempt to formalise the performance stance on understanding, i.e. we will try to define $X$'s \textit{information utilisation} of $p$. To do this, we will need to construct a set of `fair tests of $p$' for $X$. 
We will start by noting: (1) A fair test for $X$ should \textit{require} $X$'s background information; (2) a test of $p$ should \textit{require} information about $p$. We will use the formalisation of rational decision-making, AIXI, to `benchmark' how information is utilised in a given environment. Unlike a typical test-taker, AIXI enters into an environment with no prior knowledge, and thus we must present any priors to AIXI as a part of its percept sequence. Therefore, to decide whether or not a given task meets the criteria outlined above we will construct a `meta-task' for AIXI where relevant observations are prepended to the task. 

\begin{definition}[$(X,p)$-tests] \label{def:x_p_tests}
    Given a phenomenon $p \in \mathcal{P}$ and an agent $X$ with internal state $z_X \in \mathcal{Z}_X$, we start with the set of bounded-test environments $E$, we define the set of $(X,p)$-tests, $E_{X,p}$, as follows:
    \begin{align}
        E_{X,p} = \Set{
            \mu \in E ~:~ 
            V_{(p, b) \circ \mu}^{AIXI} = V_\mu^* > 0,~ 
            V_{(p) \circ \mu}^{AIXI} = 
            V_{(b) \circ \mu}^{AIXI} = 
            V_{\mu}^{AIXI} = 0
        }
    \end{align}
    Where $b$ is a shorthand for the $p$-background information $b=z_{X, *\neg p}$, and $\vec{x} \circ \mu$ denotes the construction of a new environment $\mu'$ such that:
    \begin{align}
        &\forall x_i \in \vec{x},~ \forall a_{<i},~ \mu'\brac*{(x_i, 0) ~|~ a_{<i}} = 1 \\
        &\forall k > |\mathbf{x}|,~ \mu'(e_k ~|~ a_{<k}) = \mu(e_k ~|~ a_{j\ldots k}), ~\text{where}~ j = |\mathbf{x}|
    \end{align}
\end{definition}

It is worth noting why we are using only the $p$-background information and not the agent's entire internal state as required prior knowledge. This is because if the agent knows anything about $p$ then AIXI would be able to use the information encoded in the internal state to pass the test when only given $b$. We want AIXI to only get information about $p$ from $p$ itself so that we can strictly outline the criteria above.

Using the set of fair tests for $X$, we can define a measure of \textit{information utilisation} by measuring the agent's intelligence across these environments. This is an adaptation of \citepos{Hutter2005UniversalIntelligence} measure of intelligence (Equation \ref{eqn:universal_intelligence_measure}).

\begin{definition}[$p$-Utilisation]
    Given an agent $X$ and phenomenon $p$, the $p$-utilisation $\Upsilon_p: \mathcal{X} \rightarrow [0, 1]$ is defined:
    \begin{align}
        \Upsilon_p(X) = \sum_{\mu \in E_{X,p}} 2^{-K(\mu)} V_\mu^X
    \end{align}
\end{definition}

\subsection{Information Integration}

With the definitions we have constructed here, we can also introduce a measure of how `integrated' the $p$-relevant information is.

\begin{definition}[$p$-Integration]
    Suppose we have a phenomenon $p \in \mathcal{P}$, and an agent $X \in \mathcal{X}$ with $p$-relevant information $z_{X,*p}$ and $p$-specific information $z_{X,p}$. The $p$-integration, $\phi: \mathcal{X} \times \mathcal{P} \rightarrow [0, 1)$, is defined,
    \begin{align}
        \phi(X, p) = \tanh\brac*{\frac{|\langle z_{X,*p} \rangle_X |}{|\langle z_{X,p}\rangle_X|} - 1}
    \end{align}
\end{definition}

As the $p$-relevant information will always be larger-than or equal to the $p$-specific information ($|\langle z_{X,*p} \rangle_X | \geq |\langle z_{X,p}\rangle_X|$), the ratio in this measure will equal 1 when all relevant information is specific. In this case, there is no relevant information that is used for anything else, i.e. the $p$-relevant information is not at all integrated into the rest of the internal state (or nothing else exists to integrate with). Conversely, the smaller the specific information gets, the more the relevant information must be sharing with encodings for other phenomena. 

\subsection{The Measure of Understanding}

Finally we bring these ideas together to define our measure of understanding. The resulting measure is bounded by 0 and 1.

\begin{definition}[Understanding]
    Given an agent $X \in \mathcal{X}$ with internal state $z_X$ and phenomenon $p \in \mathcal{P}$, the measure of $X$'s \textbf{understanding} of phenomenon $p$, $\understanding: \mathcal{X} \times \mathcal{P} \rightarrow [0, 1)$, is defined as:
    \begin{align}
        \understanding(X, p) &=  \frac{\hat{\understanding}(X, p) \cdot c(X, p) \cdot \phi(X, p) \cdot \Upsilon_p(X) \cdot I(z_X;p)}{K(p)}
    \end{align}
    Where $\hat{\understanding}(X, p) \in \{0, 1\}$ is $X$'s self-reported understanding of $p$.
\end{definition}

\section{Explanation Games}

With our measure of understanding, we are ready to define explanatory effectiveness:

\begin{definition}[Explanatory Effectiveness]
    The \textbf{effectiveness} of an explanation is the change in explainee's understanding of the explanandum $p \in \mathcal{P}$ over the course of the explanatory process. Formally, given an explainer agent $A$ and an explainee agent $B$ that interact over $\tau$ time steps, the explanatory effectiveness is a function $\effectiveness: \mathcal{O}_B^* \times \mathcal{P} \rightarrow (-1, 1)$ defined as:
    \begin{equation}\label{eqn:effectiveness}
        \effectiveness(\mathbf{o}_B,~p) = \understanding(B^\tau,~ p) - \understanding(B^1,~ p)
    \end{equation}
    Where $B^t$ denotes $B$ at time $t$ and $\mathbf{o}_B$ is the sequence of observations that $B$ made during the interaction.
\end{definition}

\begin{definition}[Explanation Game]
    Suppose an explainer agent $A$, explainee agent $B$, and explanandum $p \in \mathcal{P}$. An \textbf{explanation game} $G=(A, B, p, \tau)$ is a cooperative finite sequential game with asymmetric information in which the participants seek to maximise $\effectiveness(\mathbf{o}_B, p)$ over the course of $\tau$ time steps, where $\mathbf{o}_B$ is the sequence of all observations made by $B$.
\end{definition}

From these definitions, there are a few observations that we can make. Firstly, there is nothing to stop a game from having negative effectiveness, i.e. the explainee understands the phenomenon less after the `explanation'. As $\understanding$ is bounded by 0 and 1, $\effectiveness$ is bounded by -1 and 1. Secondly, there is no necessary link between effectiveness and the explainee's beliefs regarding their own understanding. It is possible for the explainee to believe that the explanation was more effective than it was (e.g. $\hat{\understanding}(X, p) = 1$, but $I(z_X; p) = 0$). Thirdly, we can use this notion to discuss the motivation of the explainer. It makes sense to consider an agent as an explainer, rather than a deceiver, only if they expect the sign of the $\effectiveness$ to be positive. Finally, it is worth noting that this measure changes according to time in which we choose to record it. The explainer may start out strong and increase the explainee's understanding of the explanandum, but then say something that leads to confusion.

\section{Discussion}

In this paper we have presented a formal model for assessing the the `explanatory effectiveness' $\effectiveness$ of a dialectical process between two agents. We used this to define \textit{explanation games} in which participants seek to maximise $\effectiveness$. Along the way we used AIT and UIT to develop a measure of an agent's `understanding' of a given phenomenon $p$. This involved partitioning the information in the agent's mental state into four objects relative to $p$; the $p$-relevant, $p$-irrelevant, $p$-specific, and $p$-background information. We used these to define the $p$-compression factor (how compressed the agent's representation of $p$ is), $p$-integration (what proportion of the representation is only encoding for $p$), and the $p$-utilisation. For the last of these we needed to construct a set of `fair tests', i.e. a set of environments that would rely on both knowledge of $p$ and the agent's background knowledge to solve. We find these environments by asking: ``Could AIXI solve this environment when given this information?''. However, it should not be taken for granted that this is the right question to ask, and thus we should study this space of environments more precisely to see if it includes unfair tests or leaves out potential fair tests. 

Future work should investigate the trustworthiness of explanations generated in our framework, as we have made the implicit assumption that if an agent understands something they can assess whether or not they trust it. One direction to look in is the implications of explainees with limited capacities, i.e. either time/space complexity constraints, or explainees who are biased in particular ways. Additionally, the assumption that explanation games are always cooperative should be challenged, as in many real situations participants may have conflicting or ulterior agendas.  For both the cooperative and non-cooperative case a useful research project will be to articulate rules for the dialogue game between explainer and explainee \citep{McBurney2002GamesAgents} and to develop strategies for each player, given their goals.  Finally, as $K$ and AIXI are not computable, alternatives for these components (for the purposes of this framework) should be devised and studied. 

\bibliography{references}   


\end{document}